# Machine learning approach to dynamic risk modeling of mortality in COVID-19: a UK Biobank study

Mohammad A. Dabbah [1], Angus B. Reed [1], Adam T.C. Booth [1], Arrash Yassaee [1,2], Alex Despotovic [1,3], Benjamin Klasmer [1], Emily Binning [1], Mert Aral [1], David Plans* [1,4], Alain B. Labrique [5], Diwakar Mohan [5]

1. Huma Therapeutics Limited, London, United Kingdom
2. Centre for Paediatrics and Child Health, Faculty of Medicine, Imperial College London, London, United Kingdom
3. Faculty of Medicine, University of Belgrade, Belgrade, Serbia
4. University of Exeter, SITE, Exeter, United Kingdom
5. Johns Hopkins Bloomberg School Public Health, Baltimore, Maryland, United States

* **Corresponding Author:** David Plans (david.plans@huma.com)

# Abstract

The COVID-19 pandemic has created an urgent need for robust, scalable monitoring tools supporting stratification of high-risk patients. This research aims to develop and validate prediction models, using the UK Biobank, to estimate COVID-19 mortality risk in confirmed cases. From the 11,245 participants testing positive for COVID-19, we develop a data-driven random forest classification model with excellent performance (AUC: 0.91), using baseline characteristics, pre-existing conditions, symptoms, and vital signs, such that the score could dynamically assess mortality risk with disease deterioration. We also identify several significant novel predictors of COVID-19 mortality with equivalent or greater predictive value than established high-risk comorbidities, such as detailed anthropometrics and prior acute kidney failure, urinary tract infection, and pneumonias. The model design and feature selection enables utility in outpatient settings. Possible applications include supporting individual-level risk profiling and monitoring disease progression across patients with COVID-19 at-scale, especially in hospital-at-home settings.

# Introduction

The COVID-19 pandemic has posed a significant challenge to global healthcare systems. Although large-scale vaccination programmes have begun, many countries will not have widespread access to vaccines until 2023, meaning that non-pharmaceutical interventions are likely to remain indispensable national strategies for some time[1].

COVID-19 shows highly varied clinical presentation, with a significant proportion (17-45%) of cases being asymptomatic and requiring no specific care[2,3]. Conversely, the case fatality rate is reported between 2–3% worldwide[4]. Between these two extremes, typical symptoms include fever, continuous cough, anosmia, and dyspnoea, which may range from requiring only self-management at home to inpatient care. Understanding which individuals are most vulnerable to severe disease, and thereby in most need of resources, is critical to limit the impact of the virus.

Decision-making at all levels requires an understanding of individuals' risk of severe disease. Various patient characteristics, comorbidities, and lifestyle factors have been linked to greater risk of death and/or severe illness following infection[5–7]. Once patients are infected with SARS-CoV-2, additional physiological parameters, such as symptoms and vital signs, can inform real-time prognostication[8]. Laboratory testing and imaging can also inform risk stratification for early, aggressive intervention, though this data is only accessible to hospital inpatients, who are likely to be already severely affected[9,10].

Robust, predictive models for acquisition and prognosis of COVID-19[11–16] and resource management[17,18] have been developed to support risk stratification and population management at-scale, offering important insights for organizational decision-making. For example, QCOVID is a leading COVID-19 risk model developed using primary care data from 8 million adults[11], with evidence of external validation[19]. It is currently implemented in the NHS as a clinical assessment tool[20]. However, the individual is currently overlooked, and granular, patient-specific risk-scoring could unify decision-making across all levels. Existing individualized risk scores, however, often conflate risk of COVID-19 acquisition with risk of mortality following infection[11,14], which can limit their utility in patient management.

For predictive models to achieve impact at scale, assessment of risk factors should be inexpensive and accessible to the general population, ideally without the need for specialized testing or hospital visits. Such risk prediction tools, through improved patient triage, could be used to further increase the efficiency of, and confidence in, hospital-at-home solutions, which have shown promise in reducing hospital burden throughout the pandemic[21]. Risk scores in these circumstances need to be dynamic and contemporaneous, ideally incorporating symptoms and vital sign data to maximise utility to clinical and research teams. Therefore, the primary aim of this study is to develop and validate a population-based prediction model, using a large, rich dataset and a selective, clinically informed approach, which dynamically estimates the COVID-19 mortality risk in confirmed diagnoses.

# Results

## Clinical characteristics of patients in the derivation cohort

There were 55,118 adults in the UK Biobank (UKB) tested for COVID-19. After excluding negative test results and patients without hospital records data, 11,245 adults (aged 51–85 years, mean: 66.9, SD: 8.7) were included in the analysis, of whom 640 (5.7%) had died as a result of COVID-19 (**Supplementary Figure 1**). The mean age of survivors was 66.4 years (SD: 8.6), compared to 76.0 years (SD: 5.6) for those that died. The most common pre-existing conditions in patients were hypertension (36.2%), osteoarthritis (23.3%), and asthma (13.3%) (**Table 1**).

## Leave-one-out validation

To maximise the potential of the dataset, a leave-one-out (LOO) cross-validation approach was implemented (**Figure 1C**). In this process, each data point is used as a test set while the remaining data points are used to train a Random Forest (RF) classifier using the entire feature space. This approach is the most extreme version of k-fold cross-validation and results in lower model bias and variance than the typical train/test split method. Feature importance is calculated by aggregating individual importance across all trained classifiers in the LOO experiment. The feature selection process (**Figure 1B**) ensured the combination of data-driven insights with clinical experience, shortlisting approximately 12,000 features to 64 characteristics. The shortlisted features included: 3 vital signs; 12 symptoms; 32 pre-existing clinical conditions; 5 medications and treatments; and 13 patient characteristics (**Table 1**).

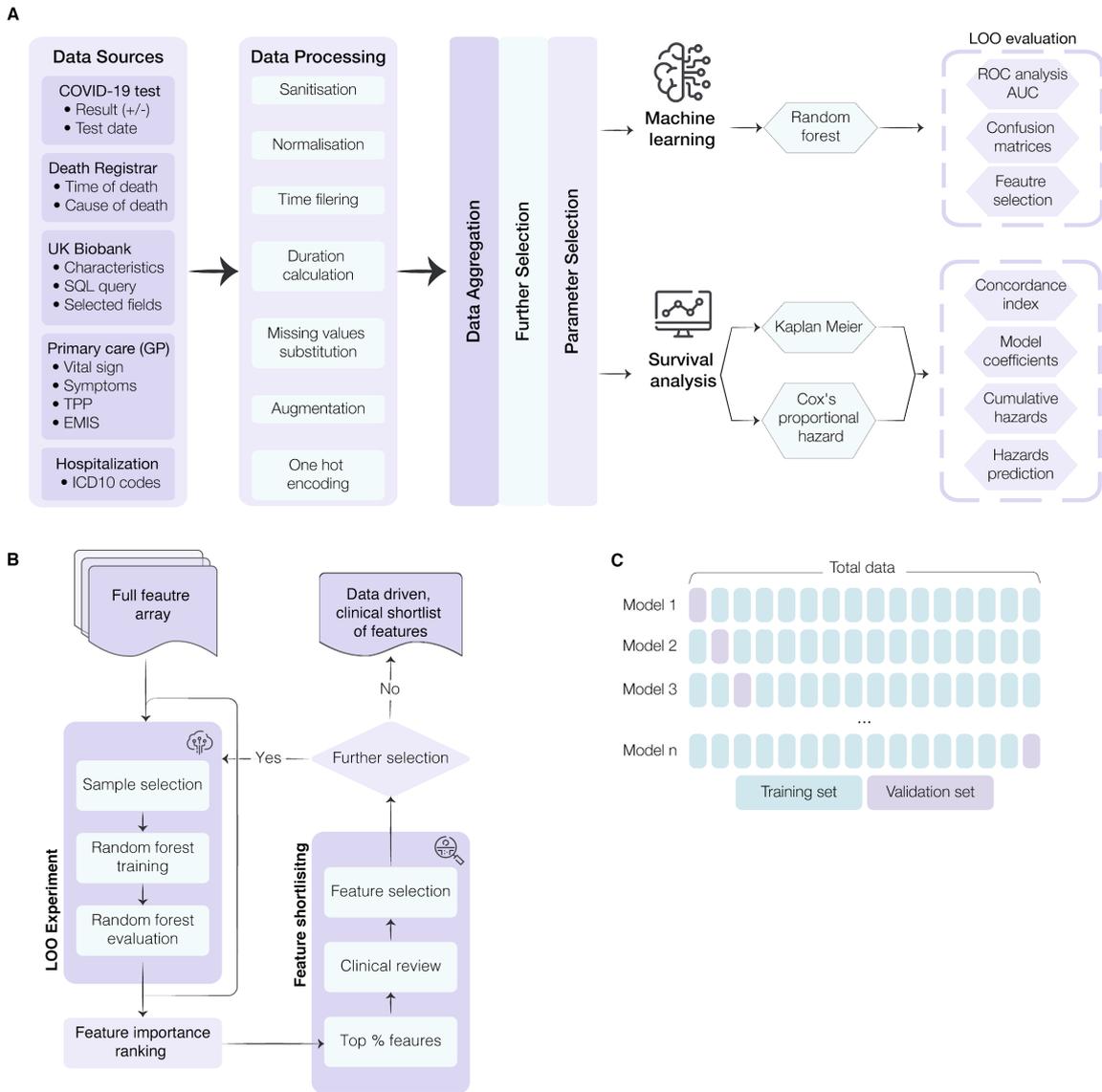

**Figure 1.** Workflow for model development and feature selection. **A)** Conceptual diagram of the data ingestion pipeline and analysis methods. To combine databases, several data pre-processing steps were carried out, including: sanitisation (eliminating redacted records and nuanced entries); normalization (scaling values to ensure fitting with a reasonable range for further processing); time filtering; duration calculation (computing the time interval between testing positive and mortality); missing value substitution (replacing missing values or records with the mean value of the UK Biobank database); augmentation (bringing all data for each subject into a single unified record); and one-hot-encoding (codifying the presence of a pre-existing condition or symptom into a binary sequence for each subject). This data ingestion process standardized the input features and attributes for all subjects in this study regardless of their unique and variable conditions, symptoms, vital signs, and records. **B)** Illustration of the data-driven and clinically reviewed feature refinement process. **C)** Schematic representation of the leave-one-out cross-validation method for feature selection and model validation. Each sample is systematically left out in each fold (purple). Prediction error estimates are based on left out samples. AUC = area under the curve; GP = general practice; LOO = Leave-One-Out; ROC = receiver operating characteristic.

## Model performance

The receiver operating characteristic (ROC) curves for the prediction models are presented in **Figure 2A.** With an area under the curve (AUC) of 0.90, the Random Forest (RF) model showed excellent performance. A Cox Proportional Hazard (CPH) model was trained using the final set of RF-defined variables to maximize explainability of the RF. This model had improved performance, reaching a higher AUC of 0.91. The results of this model highlighted both known and novel risk factors for mortality in COVID-19 (**Figure 3**). Age was the most important feature of the model. To test for overfitting due to this feature, and limitations in the dataset, the model was re-processed excluding age, which had minimal effect on model performance (CPH AUC: 0.90, **Supplementary Figure 2**).

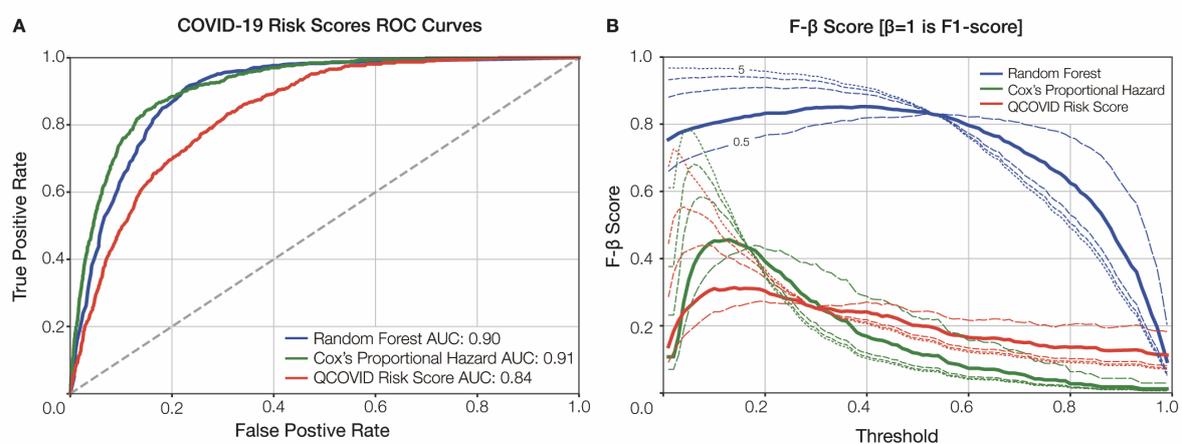

**Figure 2**. Model performance evaluation. **A)** the receiver operating characteristic (ROC) curve comparison shown for our Random Forest (RF) and Cox models against QCOVID. **B)** the F-β score generated at β=1 (F1-score in bold), β=[ 0.5, 2, 3, 5], shown in decreasing size dashed line. AUC = area under the curve. Both the ROC and F-β score curves show the performance at various thresholds (i.e. operation points). Threshold value may be dependent on the application of the model. For example, in clinical circumstances requiring low false negatives, the threshold would be optimised for recall, though this would also correspond to higher numbers of false positives.

## Novel features

Novel features highlighted by the CPH included demographic and lifestyle features, such as waist circumference and sleep duration (**Figure 3**). Key features in recent medical history are also elucidated, with prior acute kidney failure, respiratory failure, bacterial pneumonia, and non-bacterial pneumonia (diagnosed between one week and one month prior to COVID-19 infection) comprising the most prominent predictors of mortality. Acute kidney failure and bacterial pneumonia remain in the top features, even with a more distant diagnosis window of between one month and 12 months prior to COVID-19 infection (**Figure 3**).

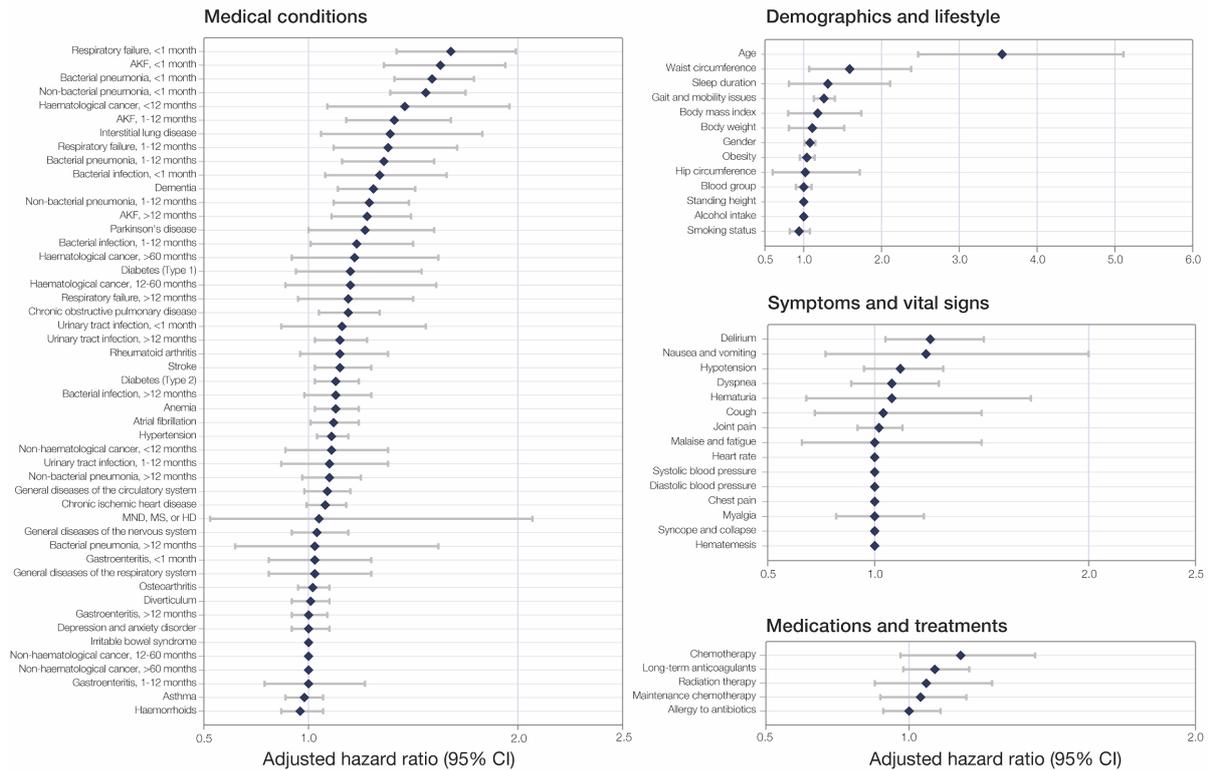

**Figure 3.** Plot of Cox model coefficients of COVID-19 mortality in UK Biobank cohort. Values show HR ± 95%CI. AKF = acute kidney failure, MND = motor neurone disease, MS = multiple sclerosis, HD = Huntington's disease, HR = hazard ratio, CI = confidence interval.

## Comparison of results

**Figure 2A** also shows the ROC curves for both the RF and CPH[22] models against the sex-aggregated QCOVID model[11]. As shown, the ROC curves for the RF and CPH are very comparable with a slight advantage for the CPH. From **Figure 2A**, it can be seen that when QCOVID is applied to the UKB dataset it performs well and achieves an AUC of 0.84, showcasing resilience to unseen data. To explore the performance further, it is essential to look at the robustness of the generated models. **Figure 2B** illustrates the use of F-β statistical analysis to examine the performance of the various models. As expected, despite the CPH having a slightly greater AUC score, it is clear that the RF has much more stable performance. Moreover, it can be seen that both the CPH and QCOVID models achieve optimal F-β scores when β is small. However, for the RF model, the F-β scores are considerably larger than its comparators and are more consistent across the range of thresholds, thereby demonstrating greater stability and increased capabilities regarding recall (i.e. minimizing false-negatives).

# Discussion

This study developed and validated machine learning models to predict mortality in patients with COVID-19 using comprehensive data from 11,245 COVID-19 patients in the UKB. The results show that by using easily accessible patient characteristics, brief medical history, symptoms, and vital signs we can predict mortality in patients with COVID-19 with excellent performance (AUC: 0.91). The features selected in the presented model mirror much of the current clinical understanding regarding risk factors associated with COVID-19 mortality, highlighting age[23] and obesity[24] as significant contributors.

In addition, we identified many novel features that may be strong predictors of mortality in patients with COVID-19. The most interesting findings concern the impact of prior inpatient diagnosis of urinary tract infection (UTI), respiratory failure, acute kidney failure, bacterial and non-bacterial pneumonias, and other bacterial infections. With the exception of UTI, dividing each feature into time groupings by their proximity to the COVID-19 diagnosis highlights diminishing risk the more distant the event. For respiratory conditions and other infections, the risk returns to approximately baseline when >12 months prior to COVID-19 diagnosis. The outlying significance of acute kidney failure at >12 months before COVID-19 diagnosis suggests the impact of renal damage may be more integral to COVID-19 prognosis than that of the respiratory system. This is supported by findings related to UTIs, where they appear as a less severe, but persistent, risk factor regardless of the time since diagnosis. Respiratory and renal complications are a hallmark of severe COVID-19[25]. It is, therefore, unsurprising that previous pathology of these organs effectively forecasts prognosis. To date, however, the relationship between non-severe urogenital pathology and COVID-19 has not been effectively assessed. A recent systematic review on urological manifestations of COVID-19 found urinary symptoms were absent from all included studies[26]. Where data has been collected, sample sizes have been too low to draw strong conclusions. Though the occurrence of *de novo* urinary symptoms has been documented without noticeable impact on prognosis[27,28], it has been previously suggested, and recently evidenced, that the presence of pre-existing urinary conditions may be associated with a poorer disease prognosis proportional to their severity[29,30]. Our investigation provides the first reliable evidence that a history of UTI is predictive of greater COVID-19 mortality risk, roughly equivalent to the predictive value of type 2 diabetes or a prior stroke (**Figure 3**). We hypothesise that the underlying nature of this association reflects the effect of poorer, possibly sub-clinical, baseline health status. While this sub-clinical deterioration previously had no noticeable impact, in the context of a highly infective, fatal pathogen such as SARS-CoV-2, even a small deterioration can take on clinical significance.

The unique value of the UKB can be attributed to its well-established, longitudinal background dataset. Encompassing non-traditional health data, including anthropometric measurements and lifestyle insights, allows for the assessment of commonly overlooked, yet easily collectable, variables to supplement the already-known clinical factors. The ability to capture a deeper phenotype of the individual prior to infection has proved integral to the model's performance, in line with other disease-specific prediction models developed on the UKB[31–33]. Notably, we identified baseline waist circumference, height, weight, and hip circumference to be valuable independent of BMI and obesity, accounting for four of the top-seven RF-ranked features (**Supplementary Figure 3**). Although the pathophysiological link between adiposity and severe COVID-19 outcomes is not fully understood, our results indicate that comprehensive body composition may provide more granular risk profiling than BMI alone[34,35]. Moreover, while baseline sleep duration has been demonstrated to be highly predictive of all-cause mortality[36], cardiovascular diseases[37], and type 2 diabetes[38], our findings mark the first instance of its significant predictive influence within COVID-19 prognosis. While we present novel features associated with severe COVID-19 outcomes, it is important to consider that correlations identified in this observational dataset often will not reflect direct causation. We encourage further investigation of these novel features, both those previously established outside of COVID-19 and those discovered *de novo*, in a prospective setting to establish the underlying pathophysiology conferring their predictive association with COVID-19 mortality.

Our model's critical component is the distinction of variables with respect to their time of onset. Classifying variables in a time-dependent fashion enables discrimination between pre-existing conditions, symptoms, and complications. This was especially important as several of our novel features are also established complications of COVID-19. Studies have emphasised the need for distinguishing pre-existing conditions from complications of COVID-19 infection and their respective impact on prognosis[39,40] but, to our knowledge, no predictive models for this disease have stratified variables in such a way. Applied in the context of patient management, and enriched by the explainability of variable time-filtering, our results could help clarify crucial aspects of patients' past medical history and their relation to predicted prognosis. Models which forecast infection risk as a component of their mortality prediction have been criticised for generalizing human behaviour, which results in underestimation of risk factors and leaves their calibration extremely vulnerable to changes in local population dynamics[41]. One strength of our model is that the risk of mortality is predicated on the assumption of a positive COVID-19 test, avoiding the associated ambiguity of multi-event prediction and enabling its use in clinical practice.

The approach taken in the development of this model is a symbiosis of machine learning and traditional statistical modeling, boosting the acceptability of the derived algorithm. From an optimisation perspective, the objective of the model is to reduce the full feature array to a minimal subgroup (**Figure 1**) while maintaining a high prediction accuracy for COVID-19 mortality. However, by significantly reducing the number of features through the data-driven approach and clinical refinement, the model also overcomes the curse of dimensionality, where beforehand the full feature array was far greater in size than the available samples, resulting in improved performance (AUC: 0.91). The results show that both the RF and CPH models are comparable in terms of accuracy. However, the RF was integral to the CPH's construction by searching through the large feature space and selecting the most important of the original ~12,000. Moreover, the RF model is more resilient to overfitting the data, and this could explain the improved F1-scores. Owing to its stability, we would recommend the RF model as the preferred implementation in clinical practice. Given the different performance characteristics of the RF and CPH models, an ensemble of the two models may be of interest for investigation to further improve stability and performance.

Several studies have reported risk models for COVID-19 mortality. In a review of prediction models for COVID-19, Wynatts et al. report all current prediction models show high risk of bias, and lack evidence from independent external validation[42]. While this model is yet to be externally validated, we have considerably larger sample size than comparable models[42] and, by implementing LOO cross-validation, our results have reduced overall variability and bias than the traditional train-test-validate method. A recent study utilising only age, minimum oxygen saturation during encounter, and health-care setting of patient encounter as features achieved comparable results (AUC: 0.91)[43], however, the intended use of this model differs from the one presented. While the model presented outperforms QCOVID (AUC: 0.91 vs. 0.84), and best efforts were made in the comparison, it cannot be considered a direct comparison. In replication of the QCOVID algorithms, variables were mapped to related fields in the UKB, however, we were unable to confirm these were fully paired. Moreover, as the UKB is not linked to GP databases in the same manner, there were some missing variables (**Supplementary Table 2**). Importantly, contrasting with our purpose of supporting patient management, QCOVID is designed for population risk stratification to aid public health decision-making, and was used to exemplify the necessity of specific model design for specific purposes.

The COVID-19 pandemic has resulted in extraordinary acceptance of digital technology in healthcare[44]. Risk assessment tools can support the streamlining of clinical time and resource prioritization, whether on a national, organizational, or patient level. Models such as those presented, can support the latter by monitoring patients at-scale and identifying those at-risk of severe illness, in real-time, and without requiring specialist equipment or clinical input. Algorithm performance may be further improved by inclusion of passive, continuous variables via smartphones or wearables. Establishing our model in a prospective healthcare setting may enable this when coupled with high quality, continuous vital sign information and replete data on the course of symptomatology. Similar digital phenotyping has also shown potential in predicting COVID-19 infection at early symptoms onset[45,46]. We believe a combination of these two types of digital tools, in union with dedicated hospital-at-home services, may become considered standard practice in infectious disease management, particularly during historically resource-intense periods, such as annual influenza outbreaks.

While the use of the UKB is a key strength in the development of the model, there are associated limitations which may impact the generalizability of the model. The UKB cohort trends towards being healthier and wealthier than the general population, which poses a notable limitation when modeling noncommunicable diseases[47]. As COVID-19 acquisition, however, is determined by exposure, this limitation is minimised in our investigation. Separately, the UKB COVID-19 data subset is less likely to capture asymptomatic or non-severe cases, in part as such individuals may not have received a test or sought medical treatment, but predominantly owing to UKB's enrichment for older age resulting in lesser rates of such presentation. The restricted age distribution (51-85 years) may further limit generalization of our findings to outside of this age range, however, Office for National Statistics figures show those aged ≥50 have accounted for 97.97% of all COVID-19-related deaths in England and Wales (up to 19$^{th}$ February 2021)[48].

Although age is clearly an important feature, our sensitivity analysis (**Supplementary Figure 2**) demonstrated negligible performance drop, likely because much of the risk associated with older age is captured within other included features. One reason for using uniform leave-one-out (LOO) training is to overcome such issues of feature reliance and generalize the model as much as possible. The F-score in **Figure 2B** illustrates this robustness, however, this must be tested on a separate representative dataset for a conclusive answer. Our robust development approach, paired with deep individual phenotyping, strengthens the evidence towards effective COVID-19 risk profiling. In addition to the limitations of the dataset, it is likely that there are regional variances in COVID-19 outcomes. As such, the model would strongly benefit from external validation, especially with the continued emergence of disruptive SARS-CoV-2 variants[49]. Evidence of real-world utility, with the associated incomplete and missing data, is lacking for COVID-19 risk models. Further research is required to both establish prospective, real-world model performance and to understand the maximal data quality reduction which still produces clinically acceptable performance.

# Conclusion

In conclusion, we present a comprehensive, robust model based on readily accessible factors (AUC: 0.91). In our analysis, we combine data-driven model development and clinical refinement to produce a model that uniquely incorporates time-to-event, symptoms, and vital signs. We identify several significant novel predictors of COVID-19 mortality with equivalent or greater predictive value than established high-risk comorbidities, such as detailed anthropometrics, lifestyle factors, and prior acute kidney failure, urinary tract infection, and pneumonias. The design and feature selection of the framework lends itself for deployment at-scale in a digital setting. Possible applications of this include supporting individual-level risk profiling and monitoring deterioration in high volumes of COVID-19 patients, particularly in hospital-at-home settings.

# Online Methods

## Study population

The development and validation of the risk model was carried out using the UKB. The UKB is a large cohort study with rich phenotype mapping of participants, including over 500,000 individuals aged between 40- and 69-years-old at recruitment, between 2006 and 2010, from across England, Scotland, and Wales[50]. The open dataset contains detailed health data and outcomes obtained prospectively from electronic health records and self-reported health measures from on-site testing over the past 15-years. The current analysis was approved under the UKB application number 55668. Ethical approval was granted by the national research ethics committee (REC 16/NW/0274) for the overall UK Biobank cohort.

## COVID-19 Status and Sample Selection

For this study, only participants with a positive RT-PCR COVID-19 test were included (**Supplementary Figure 1**). Public Health England provided data on SARS-CoV-2 tests, including the specimen date, location, and result[51]. COVID-19 test result data were available for the period 16th March 2020 to 24th February 2021, and were linked with hospital admission (28th February 2021), primary care (1st October 2020), and death records (16th February 2021). In total, 101,062 COVID-19 tests were conducted on 55,118 participants in the available cohort. Of these, 42,599 were excluded due to negative test results. Overall, 12,519 participants tested positive of which 10,605 were survivors, 640 non-survivors, and 1,274 were excluded due to missing information. Deaths were defined as COVID-19-related if ICD-10 codes U07.1 or U07.2 were present on the death records. No COVID-19 test data were available for UKB assessment centers in Scotland and Wales, thus data from these centers were not included.

## Time Filtering

Considering the chronology of medical events is critical to distinguish between, for example, pre-existing conditions and complications resulting from COVID-19. Specific attributes, therefore, can be included or excluded in the prediction model for various use cases. This study focuses on developing a model to predict mortality for COVID-19 patients before hospital admission. Accordingly, inclusion of respiratory failure (ICD-10: J96.9), for example, as a symptom or complication to predict mortality has limited use, as such events would demand hospital admission. Conversely, it is valuable to include personal history of respiratory failure as a prognostic indicator. Thus, we implemented a time filter for all features which were not demographics, symptoms, or vital signs, excluding any data recorded less than one-week prior to patients' positive COVID-19 test. This accounted for the circumstance whereby a patient may have been admitted for severe symptoms of COVID-19 prior to receiving a test. Further time filtering of <1 month, 1-12 months, and >12 months was applied to specific acute features, and <12 months, 12-60 months, and >60 months for cancer related diagnosis, to provide more granular insight. Similarly, it is important to consider only relevant symptoms and vital signs corresponding to the period of COVID-19 infection. Thus, a two-week window pre- and post- the first COVID-19 positive test was implemented.

## COVID-19 Mortality Model

### Feature Selection

The data ingestion pipeline, **Figure 1A**, generates an array of ~12,000 dimensions (including patient characteristics, pre-existing conditions, symptoms, and vital signs). Owing to the disparity in size between the survivor and non-survivors population in the dataset and the importance of obtaining an unbiased model, a LOO cross-validation experiment[52], which is also closely related to the jack-knife estimation method[53], was used to search the full feature array for the most relevant features. LOO iterates through every sample in the dataset, whereby at each step the current sample was used to evaluate the model trained on the remaining dataset (**Figure 1C**). At each iteration the samples of all classes were balanced to ensure unbiased training and, following evaluation, the model was discarded and a new model trained. A RF model was chosen due to its inherent ability to extract features, handle high dimensionality data, and generalize well to unseen data[54]. During each step of the LOO cross-validation, a ranked list of features was extracted and averaged across the entire experiment to obtain a final shortlist of features that produced the highest accuracy, further cross-checked by clinical expertise. **Figure 1B** illustrates the production of shortlisted features driven by data, and their validation and review based on clinical judgement.

Clinical feature selection was informed by a review of ranked feature importance in RF model. The highest ranked 1,000 features were screened by at least two reviewers. Any disagreements were settled by consensus with input of additional reviewers. Features were excluded where: *i*) they could not be readily obtained through self-reporting or measured outside of the clinical setting; *ii*) there was high confounding with higher ranked features; *iii*) clinical consensus concluded that the feature's rank was more likely to be explained by database bias. Subsequently, features which were closely related (e.g. cancer diagnoses) were grouped together. Supplementary ICD-10 codes were included and, where possible, generalized (**Supplementary Table 1**).

## Model Construction and Validation

The LOO evaluation was selected to maximize the value of the available datasets. The LOO is used in this case to quantitatively evaluate the model; it is not used for hyperparameter tuning of the model. In essence, at each iteration of the LOO, there is a hold-out test set, which is a single sample of unseen data. At each iteration, a completely new model is trained from scratch on a randomly selected set of samples and tested on a single hold-out sample. At the end of the experiment, following iteration over all dataset samples, the results of each of these hold-out sets are aggregated to provide the final evaluation performance of the model. None of the models at each iteration are used in any other iteration and they are completely discarded once the iteration is complete. Specifically, this is equivalent to a k-fold evaluation, where $k = n - 1$, with *n* being the total number of samples in the set. Moreover, LOO has been chosen to be as objective as possible when reporting on the outcome of the model. A single hold-out set could potentially provide a different benchmark depending on the random split of this set. Conversely, the LOO exhaustively tests against every sample in the dataset.

In this study, the prediction classes were two: COVID-19 survivors (n=11,245) and non-survivors (n=640). At each LOO iteration, two groups of equal sample size were randomly selected without replacement for training. The evaluation sample outcome and RF likelihood value were aggregated from all iterations. After aggregating all the evaluation results from the LOO experiment, the ROC curve analysis was carried out, and the AUC computed as a measure of accuracy[54]. Furthermore, the F-β statistic was used to evaluate the robustness of the model. When β is 1, this becomes the F1-score, which gives equal weights to recall and precision. A smaller β value gives more weight to precision, minimising false-positive errors, while a larger β value gives more weight to recall, minimising false-negative errors. The F-score range is [0, 1], where a score of 1 is a perfect performance.

The machine learning algorithm used in this study is the RF, which is an ensemble meta-estimator constructed from several decision trees[54]. These trees were fitted to the data using the bootstrap aggregation method (or *bagging*), which is robust and resilient to over-fitting[55]. The Gini impurity was used to compute the model likelihood of prediction. To quantify the prediction uncertainty of the RF model, a Monte Carlo approach was used to compute the confidence interval of each prediction. A CPH model[22], predicting survival time to death from the first confirmed COVID-19 positive test result, was trained on the same subset of features selected by the RF feature selection process to maximise its explainability.

## QCOVID Comparison

We compared our model against QCOVID, a leading risk prediction model for infection and subsequent death due to COVID-19, which was developed by fitting a sub-distribution hazard model on the QResearch database[11]. Predictor variables reported in QCOVID were mapped to comparable features in the UKB dataset. The UKB dataset did not include all of the relevant variables used in the QCOVID algorithm, hence chemotherapy grades and medication variables were excluded in our analysis (**Supplementary Table 2**). QCOVID risk equations for mortality were then implemented for both male and female cohorts. To ensure a fair comparison between models, QCOVID risk equations were evaluated on the UKB dataset using the same methods described above.

This article was written following the TRIPOD (Transparent Reporting of a Multivariable Prediction Model for Individual Prognosis or Diagnosis) guidelines[56], which are further elaborated in **Supplementary Table 5.**

# References

1. Reuters, Inc. China, India's COVID-19 vaccinations to stretch to late 2022: study | The Journal Pioneer. http://www.journalpioneer.com/news/world/china-indias-covid-19-vaccinations-to-stretch-to-late-2022-study-545388/.

2. Oran, D. P. & Topol, E. J. Prevalence of Asymptomatic SARS-CoV-2 Infection. *Ann. Intern. Med.* **173**, 362–367 (2020).

3. Byambasuren, O. *et al.* Estimating the extent of asymptomatic COVID-19 and its potential for community transmission: Systematic review and meta-analysis. *Off. J. Assoc. Med. Microbiol. Infect. Dis. Can.* **5**, 223–234 (2020).

4. Cao, Y., Hiyoshi, A. & Montgomery, S. COVID-19 case-fatality rate and demographic and socioeconomic influencers: worldwide spatial regression analysis based on country-level data. *BMJ Open* **10**, e043560 (2020).

5. Atkins, J. L. *et al.* Preexisting Comorbidities Predicting COVID-19 and Mortality in the UK Biobank Community Cohort. *J. Gerontol. Ser. A* **75**, 2224–2230 (2020).

6. Li, B. The Association Between Symptom Onset and Length of Hospital Stay in 2019 Novel Coronavirus Pneumonia Cases Without Epidemiological Trace. *J. Natl. Med. Assoc.* (2020) doi:10.1016/j.jnma.2020.05.015.

7. Booth, A. *et al.* Population risk factors for severe disease and mortality in COVID-19: A global systematic review and meta-analysis. *medRxiv* 2020.12.21.20248610 (2020) doi:10.1101/2020.12.21.20248610.

8. Rechtman, E., Curtin, P., Navarro, E., Nirenberg, S. & Horton, M. K. Vital signs assessed in initial clinical encounters predict COVID-19 mortality in an NYC hospital system. *Sci. Rep.* **10**, 21545 (2020).

9. Zhou F *et al.* Clinical course and risk factors for mortality of adult inpatients with COVID-19 in Wuhan, China: a retrospective cohort study. *Lancet* **395**, 1054–1062 (2020).

10. Foy, B. H. *et al.* Association of Red Blood Cell Distribution Width With Mortality Risk in Hospitalized Adults With SARS-CoV-2 Infection. *JAMA Netw. Open* **3**, e2022058 (2020).

11. Clift, A. K. *et al.* Living risk prediction algorithm (QCOVID) for risk of hospital admission and mortality from coronavirus 19 in adults: national derivation and validation cohort study. *BMJ* **371**, (2020).

12. Jin, J. *et al.* Individual and community-level risk for COVID-19 mortality in the United States. *Nat. Med.* 1–6 (2020) doi:10.1038/s41591-020-01191-8.

13. Barda, N. *et al.* Developing a COVID-19 mortality risk prediction model when individual-level data are not available. *Nat. Commun.* **11**, 4439 (2020).

14. Williamson, E. J. *et al.* Factors associated with COVID-19-related death using OpenSAFELY. *Nature* **584**, 430–436 (2020).

15. Bertsimas, D. *et al.* COVID-19 mortality risk assessment: An international multi-center study. *PLOS ONE* **15**, e0243262 (2020).

16. Yan, L. *et al.* An interpretable mortality prediction model for COVID-19 patients. *Nat. Mach. Intell.* **2**, 283–288 (2020).

17. Knight, S. R. *et al.* Risk stratification of patients admitted to hospital with covid-19 using the ISARIC WHO Clinical Characterisation Protocol: development and validation of the 4C Mortality Score. *BMJ* **370**, (2020).

18. Qian, Z., Alaa, A. M. & van der Schaar, M. CPAS: the UK's national machine learning-based hospital capacity planning system for COVID-19. *Mach. Learn.* **110**, 15–35 (2021).

19. Nafilyan, V. *et al.* An external validation of the QCovid risk prediction algorithm for risk of mortality from COVID-19 in adults: national validation cohort study in England. *medRxiv* 2021.01.22.21249968 (2021) doi:10.1101/2021.01.22.21249968.

20. Coronavirus (COVID-19) risk assessment. *NHS Digital* https://digital.nhs.uk/coronavirus/risk-assessment.

21. Shah, S. S., Gvozdanovic, A., Knight, M. & Gagnon, J. Mobile App–Based Remote Patient Monitoring in Acute Medical Conditions: Prospective Feasibility Study Exploring Digital Health Solutions on Clinical Workload During the COVID Crisis. *JMIR Form. Res.* **5**, e23190 (2021).

22. Cox, D. R. Regression Models and Life-Tables. *J. R. Stat. Soc. Ser. B Methodol.* **34**, 187–220 (1972).

23. Bonanad, C. *et al.* The Effect of Age on Mortality in Patients With COVID-19: A Meta-Analysis With 611,583 Subjects. *J. Am. Med. Dir. Assoc.* **21**, 915–918 (2020).

24. Stefan, N., Birkenfeld, A. L. & Schulze, M. B. Global pandemics interconnected — obesity, impaired metabolic health and COVID-19. *Nat. Rev. Endocrinol.* 1–15 (2021) doi:10.1038/s41574-020-00462-1.

25. Singhal, T. A Review of Coronavirus Disease-2019 (COVID-19). *Indian J. Pediatr.* **87**, 281–286 (2020).

26. Chan, V. W.-S. *et al.* A systematic review on COVID-19: urological manifestations, viral RNA detection and special considerations in urological conditions. *World J. Urol.* 1–12 (2020) doi:10.1007/s00345-020-

03246-4.

27. Dhar, N. *et al.* De Novo Urinary Symptoms Associated With COVID-19: COVID-19-Associated Cystitis. *J. Clin. Med. Res.* **12**, 681–682 (2020).

28. Mumm, J.-N. *et al.* Urinary Frequency as a Possibly Overlooked Symptom in COVID-19 Patients: Does SARS-CoV-2 Cause Viral Cystitis? *Eur. Urol.* **78**, 624–628 (2020).

29. Wu Zhang-song, Zhang Zhi-qiang, & Wu Song. Focus on the Crosstalk between COVID-19 and Urogenital Systems. *J. Urol.* **204**, 7–8 (2020).

30. Karabulut, I. *et al.* The Effect of the Presence of Lower Urinary System Symptoms on the Prognosis of COVID-19: Preliminary Results of a Prospective Study. *Urol. Int.* **104**, 853–858 (2020).

31. Alaa, A. M., Bolton, T., Angelantonio, E. D., Rudd, J. H. F. & Schaar, M. van der. Cardiovascular disease risk prediction using automated machine learning: A prospective study of 423,604 UK Biobank participants. *PLOS ONE* **14**, e0213653 (2019).

32. Rezaee, M., Putrenko, I., Takeh, A., Ganna, A. & Ingelsson, E. Development and validation of risk prediction models for multiple cardiovascular diseases and Type 2 diabetes. *PLOS ONE* **15**, e0235758 (2020).

33. Sanikini, H. *et al.* Anthropometry, body fat composition and reproductive factors and risk of oesophageal and gastric cancer by subtype and subsite in the UK Biobank cohort. *PLOS ONE* **15**, e0240413 (2020).

34. Petersen, A. *et al.* The role of visceral adiposity in the severity of COVID-19: Highlights from a unicenter cross-sectional pilot study in Germany. *Metabolism* **110**, 154317 (2020).

35. Watanabe, M. *et al.* Obesity and SARS-CoV-2: A population to safeguard. *Diabetes Metab. Res. Rev.* e3325 (2020) doi:10.1002/dmrr.3325.

36. Cappuccio, F. P., D'Elia, L., Strazzullo, P. & Miller, M. A. Sleep Duration and All-Cause Mortality: A Systematic Review and Meta-Analysis of Prospective Studies. *Sleep* **33**, 585–592 (2010).

37. Cappuccio, F. P., Cooper, D., D'Elia, L., Strazzullo, P. & Miller, M. A. Sleep duration predicts cardiovascular outcomes: a systematic review and meta-analysis of prospective studies. *Eur. Heart J.* **32**, 1484–1492 (2011).

38. Gangwisch, J. E. *et al.* Sleep Duration as a Risk Factor for Diabetes Incidence in a Large US Sample. *Sleep* **30**, 1667 (2007).

39. Guan, W.-J. *et al.* Comorbidity and its impact on 1590 patients with COVID-19 in China: a nationwide analysis. *Eur. Respir. J.* **55**, (2020).

40. Wang, Z. *et al.* Clinical symptoms, comorbidities and complications in severe and non-severe patients with COVID-19. *Medicine (Baltimore)* **99**, (2020).

41. Sperrin, M. & McMillan, B. Prediction models for covid-19 outcomes. *BMJ* **371**, m3777 (2020).

42. Wynants, L. *et al.* Prediction models for diagnosis and prognosis of covid-19: systematic review and critical appraisal. *BMJ* **369**, (2020).

43. Yadaw, A. S. *et al.* Clinical features of COVID-19 mortality: development and validation of a clinical prediction model. *Lancet Digit. Health* **2**, e516–e525 (2020).

44. Whitelaw, S., Mamas, M. A., Topol, E. & Spall, H. G. C. V. Applications of digital technology in COVID-19 pandemic planning and response. *Lancet Digit. Health* **2**, e435–e440 (2020).

45. Mishra, T. *et al.* Pre-symptomatic detection of COVID-19 from smartwatch data. *Nat. Biomed. Eng.* **4**, 1208–1220 (2020).

46. Miller, D. J. *et al.* Analyzing changes in respiratory rate to predict the risk of COVID-19 infection. *PloS One* **15**, e0243693 (2020).

47. Fry, A. *et al.* Comparison of Sociodemographic and Health-Related Characteristics of UK Biobank Participants With Those of the General Population. *Am. J. Epidemiol.* **186**, 1026–1034 (2017).

48. Office for National Statistics. Deaths registered weekly in England and Wales, provisional. https://www.ons.gov.uk/peoplepopulationandcommunity/birthsdeathsandmarriages/deaths/datasets/weekly provisionalfiguresondeathsregisteredinenglandandwales.

49. Williams, T. C. & Burgers, W. A. SARS-CoV-2 evolution and vaccines: cause for concern? *Lancet Respir. Med.* **0**, (2021).

50. Sudlow, C. *et al.* UK Biobank: An Open Access Resource for Identifying the Causes of a Wide Range of Complex Diseases of Middle and Old Age. *PLOS Med.* **12**, e1001779 (2015).

51. Armstrong, J. *et al.* Dynamic linkage of COVID-19 test results between Public Health England's Second Generation Surveillance System and UK Biobank. *Microb. Genomics* **6**, e000397 (2020).

52. Webb, G. I. *et al.* Leave-One-Out Cross-Validation. in *Encyclopedia of Machine Learning* (eds. Sammut, C. & Webb, G. I.) 600–601 (Springer US, 2011). doi:10.1007/978-0-387-30164-8_469.

53. Efron, B. *The jackknife, the bootstrap, and other resampling plans*. (Society for Industrial and Applied Mathematics, 1982).

54. Breiman, L. Random Forests. *Mach. Learn.* **45**, 5–32 (2001).

55. Breiman, L. Bagging Predictors. *Mach. Learn.* **24**, 123–140 (1996).

56. Collins, G. S., Reitsma, J. B., Altman, D. G. & Moons, K. G. Transparent reporting of a multivariable prediction model for individual prognosis or diagnosis (TRIPOD): the TRIPOD Statement. *BMC Med.* **13**, 1 (2015).

# Acknowledgements

The authors would like to thank Davide Morelli for contributions in model development. This research has been conducted using data from UK Biobank, a major biomedical database (www.ukbiobank.ac.uk).

# Author contributions

All authors have approved the final version of the manuscript submitted. All authors agree to be accountable for all aspects of the work in ensuring that questions related to the accuracy or integrity of any part of the work are appropriately investigated and resolved. M.A.D. conceived and designed the study, interpreted the results, developed the computation models, analysed the data, and wrote and reviewed the manuscript. A.B.R. and A.T.C.B. conceived and designed the study, interpreted the results, and wrote and reviewed the manuscript. B.K. interpreted the results, developed the computation models, analysed the data, and wrote and reviewed the manuscript. A.Y. and A.D. interpreted the results, and wrote and reviewed the manuscript. E.B. and M.A. conceived and designed the study and reviewed the manuscript. D.M., A.L., and D.P. interpreted the results and reviewed the manuscript.

# Disclosure statement

M.A.D., A.B.R., A.T.C.B., A.Y., A.D., B.K., E.B., M.A. and D.P. are employees of Huma Therapeutics Ltd. D.M. & A.L. declare that they have no conflicts of interest to report.

# Funding

This research was funded by Huma Therapeutics Ltd.

# Tables

| Characteristic | | All Participants | n (%) [count] Survived | Died |
|---|---|---|---|---|
| | Total | 11,245 | 10,605 (94.3) | 640 (5.7) |
| **Demographic** | | | | |
| Male sex | | 5274 | 4,850 (92) | 424 (8) |
| Age (yrs), mean (SD) | | 66.9 (8.7) | 66.4 (8.6) | 76.0 (5.6) |
| **Lifestyle and anthropometrics** | | | | |
| Body mass index, mean (SD) | | 28.4 (5.1) [11,153] | 28.3 (5.1) [10,528] | 30.0 (5.7) [625] |
| Waist circumference (cm), mean (SD) | | 92.5 (14.0) [11,185] | 92.1 (13.9) [10,556] | 100.1 (14.7) [629] |
| Hip circumference (cm), mean (SD) | | 104.8 (9.9) [11,181] | 104.6 (9.8) [10,552] | 106.7 (11.3) [629] |
| Body weight (kg), mean (SD) | | 80.9 (16.9) [11,172] | 80.6 (16.7) [10,544] | 85.9 (18.6) [628] |
| Obesity (BMI > 30) | | 1307 | 1167 (89.3) | 140 (10.7) |
| Standing height (cm), mean (SD) | | 168.5 (9.2) [11,245] | 168.5 (9.2) [10,605] | 168.9 (9.3) [640] |
| Blood type | | | | |
|   Unknown | | 353 | 318 (90.1) | 35 (9.9) |
|   AA | | 892 | 834 (93.5) | 58 (6.5) |
|   AB | | 435 | 417 (95.9) | 18 (4.1) |
|   AO | | 4,074 | 3,858 (94.7) | 216 (5.3) |
|   BB | | 67 | 62 (92.5) | 5 (7.5) |
|   BO | | 1,051 | 999 (95.1) | 52 (4.9) |
|   OO | | 4,373 | 4,117 (94.1) | 256 (5.9) |
| Sleep duration (hrs), mean (SD) | | 7.0 (1.4) [11,245] | 7.0 (1.4) [10,605] | 7.2 (1.7) [640] |
| Alcohol intake | | | | |
|   Unknown | | 33 | 30 (90.9) | 3 (9.1) |
|   Daily or almost daily | | 1,662 | 1,562 (94) | 100 (6) |
|   Three or four times a week | | 2,168 | 2,068 (95.4) | 100 (4.6) |
|   Once or twice a week | | 3,010 | 2,862 (95.1) | 148 (4.9) |
|   One to three times a month | | 1,284 | 1,228 (95.6) | 56 (4.4) |
|   Special occasions only | | 1,398 | 1,300 (93) | 98 (7) |
|   Never | | 1,690 | 1,555 (92) | 135 (8) |
| Smoking status | | | | |
|   Unknown | | 68 | 60 (88.2) | 8 (11.8) |
|   Never | | 6,195 | 5,915 (95.5) | 280 (4.5) |
|   Previous | | 3,933 | 3,642 (92.6) | 291 (7.4) |
|   Current | | 1,049 | 988 (94.2) | 61 (5.8) |
| Gait and mobility issues | | 68 | 60 (88.2) | 8 (11.8) |
| **Medication and treatment** | | | | |
| Allergy to antibiotics | | 1,143 | 1,044 (91.3) | 99 (8.7) |
| Long-term use of anticoagulants | | 981 | 821 (83.7) | 160 (16.3) |
| Radiation therapy | | 274 | 237 (86.5) | 37 (13.5) |
| Maintenance chemotherapy | | 476 | 420 (88.2) | 56 (11.8) |
| Chemotherapy | | 256 | 210 (82) | 46 (18) |
| **Pre-existing medical conditions** | | | | |
| General diseases of the circulatory system | | 1,216 | 1,030 (84.7) | 186 (15.3) |
| Chronic ischemic heart disease | | 1,388 | 1,200 (86.5) | 188 (13.5) |
| Atrial fibrillation | | 1,007 | 834 (82.8) | 173 (17.2) |
| Hypertension | | 4,074 | 3,624 (89) | 450 (11) |
| Stroke | | 767 | 624 (81.4) | 143 (18.6) |
| General diseases of the respiratory system | | 169 | 143 (84.6) | 26 (15.4) |
| Asthma | | 1,497 | 1,391 (92.9) | 106 (7.1) |
| Chronic obstructive pulmonary disease | | 670 | 537 (80.1) | 133 (19.9) |
| Interstitial lung disease | | 107 | 71 (66.4) | 36 (33.6) |
| Respiratory failure | | | | |
|   less than 1 month | | 291 | 171 (58.8) | 120 (41.2) |
|   between 1 and 12 months | | 180 | 117 (65) | 63 (35) |
|   more than 12 months | | 154 | 109 (70.8) | 45 (29.2) |
| Non-bacterial pneumonia | | | | |
|   less than 1 month | | 812 | 542 (66.7) | 270 (33.3) |
|   between 1 and 12 months | | 512 | 368 (71.9) | 144 (28.1) |
|   more than 12 months | | 624 | 508 (81.4) | 116 (18.6) |
| Bacterial pneumonia | | | | |
|   less than 1 month | | 734 | 485 (66.1) | 249 (33.9) |
|   between 1 and 12 months | | 349 | 240 (68.8) | 109 (31.2) |
|   more than 12 months | | 45 | 38 (84.4) | 7 (15.6) |
| General diseases of the nervous system | | 640 | 554 (86.6) | 86 (13.4) |
| Parkinson's disease | | 164 | 124 (75.6) | 40 (24.4) |
| MND, MS, or HD | | 21 | 18 (85.7) | 3 (14.3) |
| Dementia | | 491 | 373 (76) | 118 (24) |
| Haematological Cancer | | | | |
|   less than 12 months | | 85 | 52 (61.2) | 33 (38.8) |
|   between 12 and 60 months | | 95 | 71 (74.7) | 24 (25.3) |
|   more than 60 months | | 111 | 86 (77.5) | 25 (22.5) |
| Non-haematological Cancer | | | | |
|   less than 12 months | | 208 | 180 (86.5) | 28 (13.5) |

| | | | |
|---|---|---|---|
| between 12 and 60 months | 590 | 545 (92.4) | 45 (7.6) |
| more than 60 months | 908 | 834 (91.9) | 74 (8.1) |
| Diabetes (Type 1) | 143 | 110 (76.9) | 33 (23.1) |
| Diabetes (Type 2) | 1,416 | 1,204 (85) | 212 (15) |
| Osteoarthritis | 2,625 | 2,394 (91.2) | 231 (8.8) |
| Depression and anxiety disorder | 1,404 | 1,271 (90.5) | 133 (9.5) |
| Rheumatoid arthritis | 317 | 268 (84.5) | 49 (15.5) |
| Anemia | 1,260 | 1,067 (84.7) | 193 (15.3) |
| Urinary tract infection | | | |
| less than 1 month | 96 | 72 (75) | 24 (25) |
| between 1 and 12 months | 171 | 136 (79.5) | 35 (20.5) |
| more than 12 months | 875 | 730 (83.4) | 145 (16.6) |
| Acute kidney failure | | | |
| less than 1 month | 262 | 164 (62.6) | 98 (37.4) |
| between 1 and 12 months | 288 | 199 (69.1) | 89 (30.9) |
| more than 12 months | 443 | 331 (74.7) | 112 (25.3) |
| Any bacterial infection | | | |
| less than 1 month | 169 | 110 (65.1) | 59 (34.9) |
| between 1 and 12 months | 209 | 145 (69.4) | 64 (30.6) |
| more than 12 months | 484 | 395 (81.6) | 89 (18.4) |
| Diverticulum | 1,657 | 1,507 (90.9) | 150 (9.1) |
| Haemorrhoids | 1,120 | 1,065 (95.1) | 55 (4.9) |
| Irritable bowel syndrome | 399 | 368 (92.2) | 31 (7.8) |
| Gastroenteritis | | | |
| less than 1 month | 161 | 135 (83.9) | 26 (16.1) |
| between 1 and 12 months | 157 | 133 (84.7) | 24 (15.3) |
| more than 12 months | 1,700 | 1,546 (90.9) | 154 (9.1) |
| **Symptoms** | | | |
| Joint pain | 1,156 | 1,035 (89.5) | 121 (10.5) |
| Delirium | 250 | 175 (70) | 75 (30) |
| Hematemesis | 563 | 512 (90.9) | 51 (9.1) |
| Syncope and collapse | 19 | 17 (89.5) | 2 (10.5) |
| Dyspnea | 282 | 246 (87.2) | 36 (12.8) |
| Cough | 70 | 60 (85.7) | 10 (14.3) |
| Myalgia | 248 | 221 (89.1) | 27 (10.9) |
| Nausea and vomiting | 38 | 29 (76.3) | 9 (23.7) |
| Chest pain | 831 | 757 (91.1) | 74 (8.9) |
| Hematuria | 42 | 35 (83.3) | 7 (16.7) |
| Malaise and fatigue | 49 | 41 (83.7) | 8 (16.3) |
| Hypotension | 342 | 266 (77.8) | 76 (22.2) |
| **Vital signs** | | | |
| Diastolic blood pressure, mean (SD) | 77.9 (12.2) [123] | 77.2 (10.9) [104] | 81.9 (17.4) [19] |
| Systolic blood pressure, mean (SD) | 129.3 (19.2) [124] | 128.2 (17.6) [104] | 135.1 (25.7) [20] |
| Heart rate, mean (SD) | 84.7 (17.5) [80] | 84.0 (16.9) [71] | 90.9 (22.0) [9] |
| Body temperature, mean (SD) * | 37.5 (1.2) [41] | 37.7 (1.1) [37] | 36.1 (0.9) [4] |
| Oxygen saturation, mean (SD) * | 94.7 (3.3) [20] | 94.4 (3.6) [16] | 95.8 (1.5) [4] |
| Respiratory rate, mean (SD) * | 24.1 (7.4) [18] | 24.8 (8.5) [11] | 22.9 (5.8) [7] |

**Table 1.** Descriptive characteristics of the UK Biobank cohort with positive COVID-19 test results. Pre-existing medical conditions included only when reported more than one week prior to COVID-19 positive test result. Symptoms and vitals included only from primary care (GP) records when reported within +/- two weeks of COVID-19 positive test result. MND = motor neurone disease; MS = multiple sclerosis; HD = Huntington's disease. * Oxygen saturation, respiratory rate, and body temperature were included in the initial analysis, however, they were removed from the model due to low data availability.

**[END OF MANUSCRIPT]**